\PassOptionsToPackage{unicode}{hyperref}
\PassOptionsToPackage{hyphens}{url}
\PassOptionsToPackage{dvipsnames,svgnames,x11names}{xcolor}
\documentclass[
]{interact}

\usepackage{amsmath,amssymb}
\usepackage{iftex}
\ifPDFTeX
  \usepackage[T1]{fontenc}
  \usepackage[utf8]{inputenc}
  \usepackage{textcomp} % provide euro and other symbols
\else % if luatex or xetex
  \usepackage{unicode-math}
  \defaultfontfeatures{Scale=MatchLowercase}
  \defaultfontfeatures[\rmfamily]{Ligatures=TeX,Scale=1}
\fi
\usepackage{lmodern}
\ifPDFTeX\else
\fi
\IfFileExists{upquote.sty}{\usepackage{upquote}}{}
\IfFileExists{microtype.sty}{% use microtype if available
  \usepackage[]{microtype}
  \UseMicrotypeSet[protrusion]{basicmath} % disable protrusion for tt fonts
}{}
\makeatletter
\@ifundefined{KOMAClassName}{% if non-KOMA class
  \IfFileExists{parskip.sty}{%
    \usepackage{parskip}
  }{% else
    \setlength{\parindent}{0pt}
    \setlength{\parskip}{6pt plus 2pt minus 1pt}}
}{% if KOMA class
  \KOMAoptions{parskip=half}}
\makeatother
\usepackage{xcolor}
\makeatletter
\ifx\paragraph\undefined\else
  \let\oldparagraph\paragraph
  \renewcommand{\paragraph}{
    \@ifstar
      \xxxParagraphStar
      \xxxParagraphNoStar
  }
  \newcommand{\xxxParagraphStar}[1]{\oldparagraph*{#1}\mbox{}}
  \newcommand{\xxxParagraphNoStar}[1]{\oldparagraph{#1}\mbox{}}
\fi
\ifx\subparagraph\undefined\else
  \let\oldsubparagraph\subparagraph
  \renewcommand{\subparagraph}{
    \@ifstar
      \xxxSubParagraphStar
      \xxxSubParagraphNoStar
  }
  \newcommand{\xxxSubParagraphStar}[1]{\oldsubparagraph*{#1}\mbox{}}
  \newcommand{\xxxSubParagraphNoStar}[1]{\oldsubparagraph{#1}\mbox{}}
\fi
\makeatother

\providecommand{\tightlist}{%
  \setlength{\itemsep}{0pt}\setlength{\parskip}{0pt}}\usepackage{longtable,booktabs,array}
\usepackage{calc} % for calculating minipage widths
\usepackage{etoolbox}
\makeatletter
\patchcmd\longtable{\par}{\if@noskipsec\mbox{}\fi\par}{}{}
\makeatother
\IfFileExists{footnotehyper.sty}{\usepackage{footnotehyper}}{\usepackage{footnote}}
\makesavenoteenv{longtable}
\usepackage{graphicx}
\makeatletter
\newsavebox\pandoc@box
\newcommand*\pandocbounded[1]{% scales image to fit in text height/width
  \sbox\pandoc@box{#1}%
  \Gscale@div\@tempa{\textheight}{\dimexpr\ht\pandoc@box+\dp\pandoc@box\relax}%
  \Gscale@div\@tempb{\linewidth}{\wd\pandoc@box}%
  \ifdim\@tempb\p@<\@tempa\p@\let\@tempa\@tempb\fi% select the smaller of both
  \ifdim\@tempa\p@<\p@\scalebox{\@tempa}{\usebox\pandoc@box}%
  \else\usebox{\pandoc@box}%
  \fi%
}
\def\fps@figure{htbp}
\makeatother
\NewDocumentCommand\citeproctext{}{}

\makeatletter
 \let\@cite@ofmt\@firstofone
 \def\@biblabel#1{}
 \def\@cite#1#2{{#1\if@tempswa , #2\fi}}
\makeatother
\newlength{\cslhangindent}
\newlength{\csllabelwidth}
\newenvironment{CSLReferences}[2] % #1 hanging-indent, #2 entry-spacing
 {\begin{list}{}{%
  \setlength{\itemindent}{0pt}
  \setlength{\leftmargin}{0pt}
  \setlength{\parsep}{0pt}
  \ifodd #1
   \setlength{\leftmargin}{\cslhangindent}
   \setlength{\itemindent}{-1\cslhangindent}
  \fi
  \setlength{\itemsep}{#2\baselineskip}}}
 {\end{list}}
\usepackage{calc}

\usepackage{orcidlink}
\makeatletter
\@ifpackageloaded{caption}{}{\usepackage{caption}}
\AtBeginDocument{%
\ifdefined\contentsname
  \renewcommand*\contentsname{Table of contents}
\else
  \newcommand\contentsname{Table of contents}
\fi
\ifdefined\listfigurename
  \renewcommand*\listfigurename{List of Figures}
\else
  \newcommand\listfigurename{List of Figures}
\fi
\ifdefined\listtablename
  \renewcommand*\listtablename{List of Tables}
\else
  \newcommand\listtablename{List of Tables}
\fi
\ifdefined\figurename
  \renewcommand*\figurename{Figure}
\else
  \newcommand\figurename{Figure}
\fi
\ifdefined\tablename
  \renewcommand*\tablename{Table}
\else
  \newcommand\tablename{Table}
\fi
}
\@ifpackageloaded{float}{}{\usepackage{float}}
\floatstyle{ruled}
\@ifundefined{c@chapter}{\newfloat{codelisting}{h}{lop}}{\newfloat{codelisting}{h}{lop}[chapter]}
\floatname{codelisting}{Listing}

\makeatother
\makeatletter
\@ifpackageloaded{caption}{}{\usepackage{caption}}
\@ifpackageloaded{subcaption}{}{\usepackage{subcaption}}
\makeatother

\usepackage{bookmark}

\IfFileExists{xurl.sty}{\usepackage{xurl}}{} % add URL line breaks if available
\hypersetup{
  pdftitle={A region-wide, multi-year set of crop field boundary labels for Africa},
  pdfauthor={Estes, L.D.; Wussah, A.; Asipunu, M.; Gathigi, M.; Kovačič, P.; Muhando, J.; Yeboah, B.V.; Addai, F.K.; Akakpo, E.S.; Allotey, M.K.; Amkoya, P.; Amponsem, E.; Donkoh, K.D.; Ha, N.; Heltzel, E.; Juma, C.; Mdawida, R.; Miroyo, A.; Mucha, J.; Mugami, J.; Mwawaza, F.; Nyarko, D.A.; Oduor, P.; Ohemeng, K.N.; Segbefia, S.I.D.; Tumbula, T.; Wambua, F.; Xeflide, G.H.; Ye, S.; Yeboah, F.},
  colorlinks=true,
  linkcolor={blue},
  filecolor={Maroon},
  citecolor={Blue},
  urlcolor={Blue},
  pdfcreator={LaTeX via pandoc}}

\title{A region-wide, multi-year set of crop field boundary labels for
Africa}
\author{Estes, L.D.$\textsuperscript{1}$, Wussah,
A.$\textsuperscript{2}$, Asipunu, M.$\textsuperscript{2}$, Gathigi,
M.$\textsuperscript{3}$, Kovačič, P.$\textsuperscript{3}$, Muhando,
J.$\textsuperscript{3}$, Yeboah, B.V.$\textsuperscript{2}$, Addai,
F.K.$\textsuperscript{2}$, Akakpo, E.S.$\textsuperscript{2}$, Allotey,
M.K.$\textsuperscript{2}$, Amkoya, P.$\textsuperscript{3}$, Amponsem,
E.$\textsuperscript{2}$, Donkoh, K.D.$\textsuperscript{2}$, Ha,
N.$\textsuperscript{1}$, Heltzel, E.$\textsuperscript{1}$, Juma,
C.$\textsuperscript{3}$, Mdawida, R.$\textsuperscript{3}$, Miroyo,
A.$\textsuperscript{3}$, Mucha, J.$\textsuperscript{3}$, Mugami,
J.$\textsuperscript{3}$, Mwawaza, F.$\textsuperscript{3}$, Nyarko,
D.A.$\textsuperscript{2}$, Oduor, P.$\textsuperscript{3}$, Ohemeng,
K.N.$\textsuperscript{2}$, Segbefia,
S.I.D.$\textsuperscript{2}$, Tumbula, T.$\textsuperscript{3}$, Wambua,
F.$\textsuperscript{3}$, Xeflide, G.H.$\textsuperscript{2}$, Ye,
S.$\textsuperscript{4}$, Yeboah, F.$\textsuperscript{2}$}

\thanks{CONTACT: Estes,
L.D.. Email: \href{mailto:lestes@clarku.edu}{\nolinkurl{lestes@clarku.edu}}. }
\begin{document}
\captionsetup{labelsep=space}
\maketitle
\textsuperscript{1} Graduate School of Geography, Clark
University, Worcester, MA, USA\\ \textsuperscript{2}  Farmerline
Ltd, Kumasi, Ghana\\ \textsuperscript{3}  Spatial
Collective, Nairobi, Kenya\\ \textsuperscript{4} Institute of
Agricultural Remote Sensing and Information Technology, College of
Environmental and Resource Sciences, Zhejiang
University, Hangzhou, China
\begin{abstract}
African agriculture is undergoing rapid transformation. Annual maps of
crop fields are key to understanding the nature of this transformation,
but such maps are currently lacking and must be developed using advanced
machine learning models trained on high resolution remote sensing
imagery. To enable the development of such models, we delineated field
boundaries in 33,746 Planet images captured between 2017 and 2023 across
the continent using a custom labeling platform with built-in procedures
for assessing and mitigating label error. We collected 42,403 labels,
including 7,204 labels arising from tasks dedicated to assessing label
quality (Class 1 labels), 32,167 from sites mapped once by a single
labeller (Class 2) and 3,032 labels from sites where 3 or more labellers
were tasked to map the same location (Class 4). Class 1 labels were used
to calculate labeller-specific quality scores, while Class 1 and 4 sites
mapped by at least 3 labellers were used to further evaluate label
uncertainty using a Bayesian risk metric. Quality metrics showed that
label quality was moderately high (0.75) for measures of total field
extent, but low regarding the number of individual fields delineated
(0.33), and the position of field edges (0.05). These values are
expected when delineating small-scale fields in 3-5 m resolution
imagery, which can be too coarse to reliably distinguish smaller fields,
particularly in dense croplands, and therefore requires substantial
labeller judgement. Nevertheless, previous work shows that such labels
can train effective field mapping models. Furthermore, this large,
probabilistic sample on its own provides valuable insight into regional
agricultural characteristics, highlighting variations in the median
field size and density. The imagery and vectorized labels along with
quality information is available for download from two public
repositories.
\end{abstract}

\section{Introduction}\label{introduction}

Africa's agricultural systems are undergoing a large-scale
transformation in response to rapid economic and population growth,
increasing urbanization, climatic change, and other factors (Brückner
2012, Searchinger et al. 2015, Henderson et al. 2017, Bullock et al.
2021). These changes include accelerating cropland expansion (Potapov et
al. 2022), an increase in medium-scale farms (Sitko and Chamberlin
2015), and expanding tree crops (Kimambo et al. 2020, Fagan et al.
2022). Understanding the nature and scale of these changes is important
for answering a number of sustainability-related questions, such as the
impact to ecological and climate systems (Searchinger et al. 2015, Davis
et al. 2020, 2023). Among several key datasets needed for tracking
agricultural changes are field boundary (i.e.~parcel) maps, which
provide the foundations for characterizing agricultural systems (Fritz
et al. 2015, Estes et al. 2022, Wang et al. 2022). As one of many
examples, data on field sizes can enable farm service providers to offer
seed and fertilizer in appropriate bundle sizes (e.g. Wussah et al.
2022). Field boundary maps help to constrain downstream models for
estimating crop types and yields (e.g. Estes et al. 2013), which are
used by service providers to assess the effectiveness of their
recommendations, and by government agencies to estimate crop supply
relative to demand in order to address food shortages (Fourie 2009).
Field boundary maps therefore provide a key ingredient for understanding
many aspects of agricultural systems.

\begin{figure}

\centering{

\pandocbounded{\includegraphics[keepaspectratio]{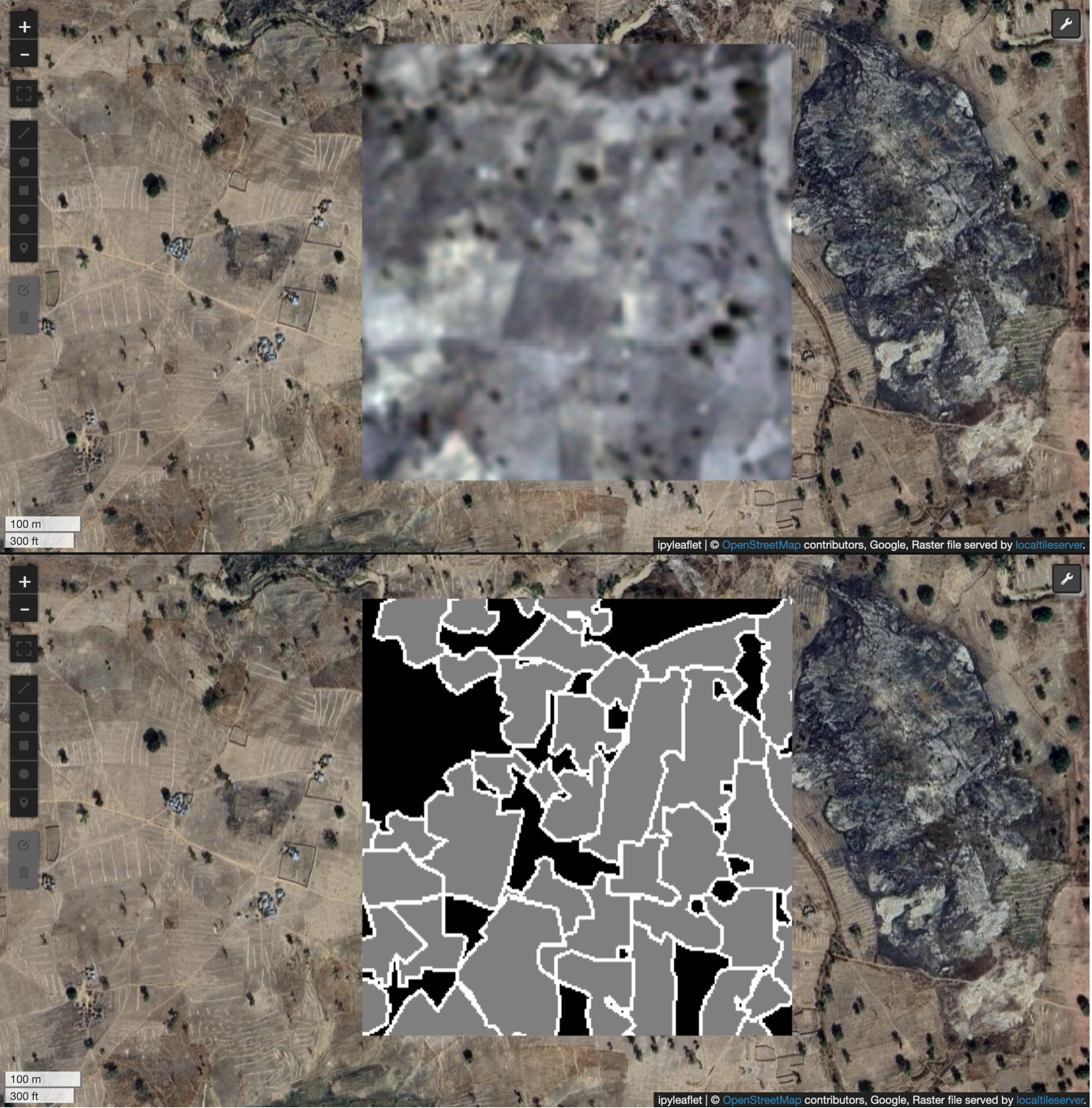}}

}

\caption{\label{fig-label}A Planet image (top) with a corresponding
image label with three classes distinguishing the non-field background
(black), field interior (grey), and field edge (white).}

\end{figure}

Field boundary maps are largely unavailable for most African countries
because of the logistical challenge and expense of collecting such data
in the field (Estes et al. 2022). The only practical tool for creating
field boundary maps is satellite remote sensing, but the small sizes and
dynamic nature of smallholder fields makes them hard to map accurately
(Estes et al. 2022, Wang et al. 2022, Rufin et al. 2024). New high
resolution satellites combined with deep learning make it increasingly
feasible to precisely map individual fields (Nakalembe and Kerner 2023),
which has been demonstrated by recent studies using very high resolution
commercial imagery (e.g. Rufin et al. 2024), as well as Planet's new
field boundary dataset\footnote{\url{https://www.planet.com/pulse/planets-newest-planetary-variable-automated-field-boundaries/}}.
While promising, these data require a subscription, therefore many
non-commercial users it will be necessary to develop their own models
and maps. Developing deep learning mapping models requires large,
geographically representative samples of labelled images (Burke et al.
2021). For field boundary mapping, models perform best when trained to
separately distinguish between field perimeters and field interiors
Waldner and Diakogiannis (2020). To train models to make these
distinctions accurately, it is necessary to carefully and precisely
digitize the boundaries of individual fields (rather than grouping
several fields into a single polygon), so that these can be converted
into rasterized labels showing field boundaries, field interiors, and
non-field areas Figure~\ref{fig-label}.

These model requirements expose a problem of label availability. Until
recently, there were no publicly available field boundary labels that
cover large, geographically diverse regions. This shortage was partially
addressed by the recent release of the Fields of the World (Kerner et
al. 2024) and GloCab (Hall et al. 2024) datasets, which both provide
labels for individual fields and corresponding Sentinel-2 imagery over a
number of countries. These two datasets will greatly enhance efforts to
train field boundary mapping models, but between them provide data for
only three African countries (Kenya, South Africa, and Rwanda in the
Fields of the World). Other relevant label sets exist that could be
combined with these to improve coverage of Africa, such as one developed
for Ghana (Estes et al. 2022), However, such data tend to represent just
one time point, making them less useful for developing multi-year maps,
given Africa's rapid agricultural changes and differences that arise
from environmental variation between different years, which combine to
limit model transferability (Khallaghi et al. 2024).

Field boundary labels that represent the continent's agricultural
systems over multiple years are therefore needed to enable the
development of models that can effectively map changes in Africa's
smallholder-dominated cropping systems. Given the cost and logistical
difficulty of ground data collection, the only practical solution is to
label directly on satellite imagery (e.g. Estes et al. 2022, Rufin et
al. 2024), which poses its own set of challenges. First,
image-interpreted labels can have substantial interpretation errors
(Estes et al. 2022). Second, high-resolution imagery (\textless5 m) is
essential for labelling, but is either expensive or hard to acquire for
a uniform time period (Lesiv et al. 2018), and temporal discrepancies
between the labelled images and those used for inference can results in
map error (Elmes et al. 2020).

To address the need for a comprehensive set of field boundary labels, we
collected a sample of more than 32,000 images throughout Africa's
croplands, spanning the years 2017-2023, and delineated the annual crop
fields visible within the imagery. We used Planet basemap mosaics that
are publicly available through Norway's International Climate and
Forests Initiative (NICFI 2024). Using these images has two major
advantages: 1) they provide cloud-free, spatially complete coverage at
monthly (from 2020) to 6-monthly (2017-2020) intervals; 2) their 4.8 m
resolution is sharp enough for humans to recognize a large proportion of
smallholder fields. Although \textless2 m imagery (very high resolution,
or VHR) is needed to most accurately distinguish smallholder fields
(Wang et al. 2022), there are no freely available sources of VHR images
that have the same temporal resolution as the Planet archive, which
provides spatially complete coverage for distinct seasons in specific
years, a crucial characteristic for tracking changing agricultural
dynamics. To address the challenge of labeling in the coarser Planet
imagery, we used a custom labelling platform designed for Planet data
with built-in procedures for assessing and mitigating label error (Estes
et al. 2016, Estes et al. 2022).

The resulting labelled images provides a dataset that will facilitate
the training and validation of field boundary mapping models, while
providing useful insight into regional variations in the characteristics
of African croplands.

\section{Methods}\label{methods}

\subsection{Sample selection}\label{sample-selection}

We collected a sample of imagery from likely cropland areas identified
using an existing moderate resolution cropland data layer developed by
the University of Maryland (Potapov et al. 2022), focusing on
continental Africa south of the Sahara where annual rainfall is above
150 mm, down to 30\(\circ\) latitude crossing the Northern Karoo in
South Africa, the southern extent of NICFI image availability. To create
the label set, we drew a random sample of \textgreater37,000 cells from
an existing \textasciitilde500 m (0.005\(^\circ\)) sampling grid (Estes
et al. 2022), which was stratified into 9 different agro-ecoregions
(ranging from Arid to Humid, as defined by the United Nations Food and
Agricultural Organization). We sampled cells that had at least 50\%
cropland cover, to ensure that most labels had a mixture of cropland and
non-cropland and to minimize the number of purely negative
(non-cropland) labels generated. The resulting cells consituted a sample
of unique mapping sites that was selected to be representative of the
region's croplands.

\subsection{Image processing}\label{image-processing}

The selected samples cells were randomly assigned to one of the 6 years
in the initially determined study timeframe (2017-2022), with the sample
subsequently adjusted to include 2023. In each year, imagery from the
least cloudy month for the given location was selected. We determined
cloudiness by calculating the monthly frequency of bad quality MODIS
pixels for the year 2022 within the Google Earth Engine platform
(Gorelick et al. 2017). We then used a set of python routines\footnote{\url{https://github.com/agroimpacts/maputil/blob/main/maputil/planet_downloader.py}}
to query the Planet API for each location and its corresponding date in
the sample. The Planet NICFI quads intersecting a larger
0.0592\(^\circ\) tile that each sample cell was located in was
downloaded, cropped, reprojected to geographic coordinates, and
resampled to a 0.000025\(^\circ\) resolution (approximately 3 m), which
enabled better visual identification of field boundaries. Within each
tile, we normalized each band within its 1st and 99th percentile ranges,
which further improved image contrast and the ability to distinguish
fields. The normalized images were converted to Cloud-Optimized
Geotiffs, and uploaded to SentinelHub\footnote{\url{https://www.sentinel-hub.com/}}
using the Bring Your Own Cog (BYOC) service, where they were accessed by
the labelling platform using the Web Map Service (WMS) protocol.

\subsection{Label collection}\label{label-collection}

The sample of sites was randomly separated into the following
categories:

\begin{itemize}
\tightlist
\item
  Class 1: Sites labelled by experts that were used for quality control
  (Q) in the labelling platform (n=2000);
\item
  Class 2: Sites to be mapped one time by one individual labeller
  (n=31,000);
\item
  Class 4: Individual samples sites mapped one time each by three
  different labeller (n=1,000).
\end{itemize}

Initially, a Class 3 sample, which represented locations with up to 6
years of imagery to be labelled, were selected, but these were
determined to be impractical to label, given the substantial variability
in image interpretation. Class 3 locations where therefore re-allocated
to Class 2 to boost that sample size.

\begin{figure}

\centering{

\pandocbounded{\includegraphics[keepaspectratio]{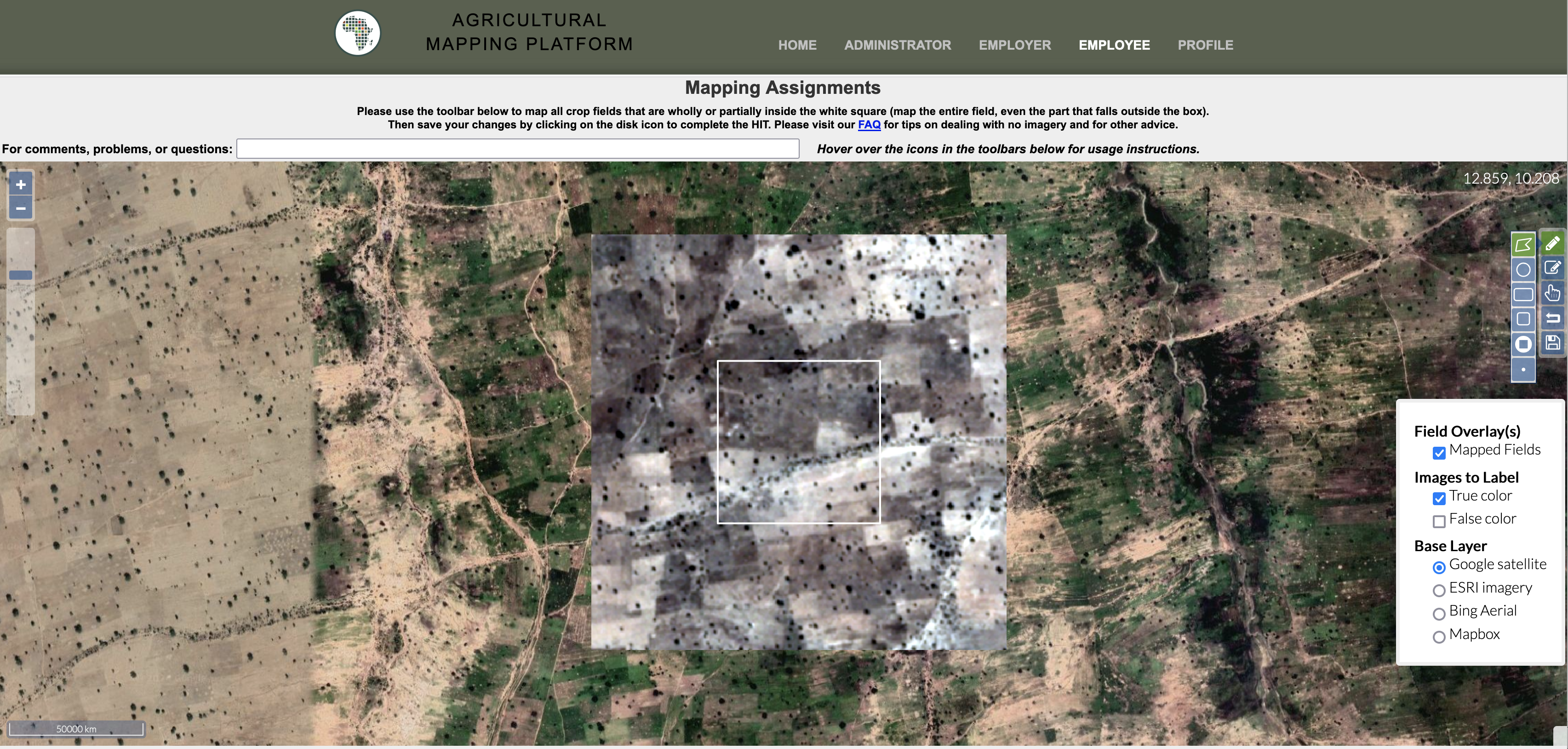}}

}

\caption{\label{fig-labeller}A view of the labelling platform's
interface, showing the target box and a true color rendering of the
Planet imagery to label at that location. Labellers were shown a larger
image extent in order to give additional context. The imagery was also
rendered in false color rendering to aid interpretation, along with
several virtual globe basemaps to provide higher resolution views.}

\end{figure}

The basic labelling task across both methods entailed mapping fields
visible in the processed Planet imagery within the target box, which is
the boundary of a cell selected from the 0.005\(^\circ\) sample grid
Figure~\ref{fig-labeller}. The objective was to delineate boundaries of
recently active fields that appeared to be under annual cultivation
(i.e.~planted to field crops) within the cell, while fields that were
determined to be inactive, abandoned, or fallow were not labelled.

Labels were collected using two methods. The primary method was through
a custom cloud-hosted labelling platform designed specifically for field
boundary delineation on Planet imagery (Estes et al. 2016, see Estes et
al. 2022). This platform was not immediately accessible, as it had to be
redesigned to work with SentinelHub's WMS services (it previously used
Element84's rasterfoundry\footnote{\url{https://www.element84.com/software-engineering/raster-foundry-geospatial-analysis-scale/}}
service). While this capability was being established, a team of expert
labellers (experienced spatial analysts) collected the Class 1 samples
using QGIS. Class 1 labels were collected by a team of 5 individual
experts using QGIS. A subset of their samples were reviewed and graded
for quality by a separate set of geospatial analysts at Clark
University, and 797 were selected to load into the labelling platform
prior to the collection of Class 2 and 4 sites. The final selected Class
1 labels were used as quality control (Q) assignments, slipped
surreptiously into each labeller's workflow, in order to measure
differences between the two sets of collected labels.

\subsection{Label quality}\label{label-quality}

The labelling task was inherently difficult, as the nature of
smallholder-dominated croplands together with the resolution of the
imagery, which combine to make discerning field boundaries difficult
under certain conditions, can often lead to situations where there are
multiple reasonable interpretations as to what constitutes a field. We
therefore used several different methods to assess and control label
quality.

\subsubsection{Training}\label{training}

The labelling teams were hired following an initial selection process in
which their ability to map a small number of sites provided in a QGIS
project was assessed. After hiring and before commencing ordinary
mapping assignments, labellers were provided with an inital round of
training by the management teams, and were required to pass a 10-site
qualification test provided on the labelling platform, in which each
person's maps were assessed against known boundaries using quality
control routines described in the next section.

\subsubsection{On platform quality
control}\label{on-platform-quality-control}

\begin{figure}

\centering{

\pandocbounded{\includegraphics[keepaspectratio]{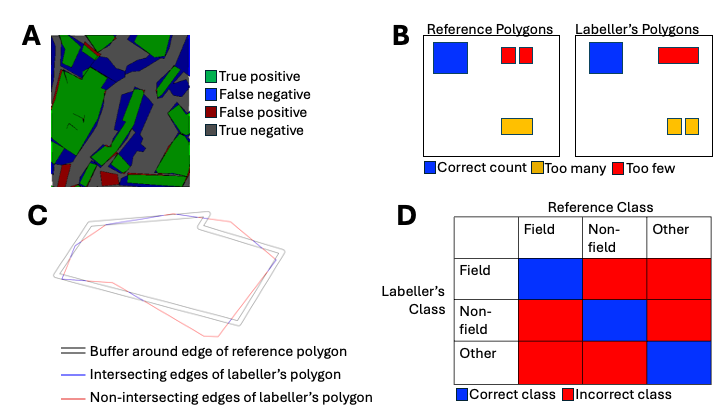}}

}

\caption{\label{fig-qmetrics}The 4 dimensions used to assess label
quality relative to Category 1 reference polygons. A = Correctly
labelled areas inside the target cell; B = Fragmentation accuracy, or
the agreement between the number of fields mapped by the reference
(Class 1) label and the labeller; C = Edge accuracy; D = Categorical
accuracy.}

\end{figure}

During labelling, each team member's work was periodically assessed
against Class 1 labels at sites that were automatically assigned by the
platform's assignment scheduling routine. The scheduler randomly
assigned different labelling tasks to individual labellers according to
a frequency parameter assigned to each label class. Quality control
tasks (Q assignments) were initially served to labellers at a rate of 1
in 10 assignments, dropping to 2 in 100 assignments towards the end of
the labelling period. Q assignments were assessed against Class 1 labels
using a multi-dimensional metric Figure~\ref{fig-qmetrics}. For this
project, 4 metrics were applied in the overall Qscore calculated after
each Q assignment was completed, which was calculated as follows:

\begin{equation}\phantomsection\label{eq-score}{
\mathrm{Qscore = 0.55Area + 0.225N + 0.1Edge + 0.125Categorical}
}\end{equation}

Here Area measures the degree of overlap in the extents digitized by the
labeller and the expert (Class 1 polygons), and is calculated as the sum
of areas of agreement for field (positive agreement) and non-field
(negative agreement) areas divided by the total area of the labelling
target (which sums positive and negative agreement and positive and
negative disagreement). N represents the agreement in the number of
fields digitized by the labeller and the expert (Class 1), while Edge is
a measure of how close the boundaries of the labeller's polygons were to
those of the Class 1 label, and Categorical is the agreement between the
class value assigned by the labeller and the expert.

\subsubsection{Label reviews}\label{label-reviews}

In addition to the on-board quality metrics assessed, a parallel review
process was undertaken by the project supervisory team. Two supervisors
conducted independent reviews of 4348 randomly selected assignments
using a standalone Jupyter Lab environment\footnote{\url{https://github.com/agroimpacts/labelreview/blob/main/review_labellers.ipynb}},
which provided a rendering of the imagery and digitized labels for each
selected site. Each site was given a score of 0-4, with 0-3 indicating
increasing levels of accuracy in assignments where fields were present
in the imagery, and 4 indicating sites that were correctly labelled as
having no fields present. A simplified binary version of this metric was
also calculated, in which ratings of 0-1 were grouped as failing, and
2-4 as passing.

To evaluate the consistency of expert ratings, we conducted an initial
assessment of between-expert agreement, in which the two supervisors'
ratings were compared for an overlapping set of 190 assignments. The two
experts agreed 46.8\% of the time along the 5 category rating system,
and 74\% of the time on the simplified binary metric. The overall mean
rating difference for the 4-class metric was 0.12 (expert 2 minus expert
1 rating), indicating no appreciable rating bias between experts.

\subsubsection{Label risk}\label{label-risk}

Additional insights into label quality can also be obtained by assessing
the differences between labels at sites mapped by multiple labellers,
which provides a measure of label uncertainty. There are multiple ways
to estimate label uncertainty, ranging from simple rasterized heat maps
to more complex, probabilistic measures that factor in label quality.
Here we illustrate label uncertainty using a Bayesian risk measure
previously applied in Estes et al. (2022), which is estimated by first
calculating a rasterized consensus label:

\begin{equation}\phantomsection\label{eq-consensus}{
C = P(\theta|D) = \sum_{i=1}^{n}P(L_i|D)P(\theta|D,L_i)
}\end{equation}\\

Here \emph{C}, consensus, is the probability (\emph{P}) that the true
value (\(\theta\)) of each pixel (field or not field) corresponds to the
most frequently labelled value (\emph{D}, field or not field) assigned
by the labellers (\emph{L}) who annotated the particular site. That
likelihood is the sum of the two multiplied probabilities on the right
hand side of Equation~\ref{eq-consensus}, which were calculated for each
labeller \emph{i}. \(P(\theta|D,L_i)\) is the labeller's label for a
given pixel. \(P(L_i|D)\) is the prior probability that the labeller was
correct when delineating labels for a specific class. It was calculated
for each labeller by multiplying their average Qscore
(Equation~\ref{eq-score}) by their average Producer's accuracy for the
selected class (see Estes et al. 2022 for full details). This formula
ensures that greater weight is given to labels produced by labellers who
were more likely to be correct. Once \emph{C} was estimated, the
consensus label \(\hat{D}\) was assigned as 0 (non-field) if \(C<0.5\),
otherwise 1 (field). Bayesian risk \emph{r} was then calculated as:

\begin{equation}\phantomsection\label{eq-risk}{
r = C(1-\hat{D}) + (1-C)\hat{D}
}\end{equation}

The maximum value of risk for either value of \(\hat{D}\) is 0.5 for any
given pixel, from which two site-wide measures of risk were assessed: 1)
the average risk; 2) the proportion of pixels exceeding a specified risk
threshold. We calculated consensus and risk from sites rasterized to
224x224 pixels, and used a risk threshold of 0.34 to calculate the
proportion of ``risky'' pixels in each consensus label.

\subsection{Cropland characteristics}\label{cropland-characteristics}

In addition to training models, these labels can also be used to provide
insights into the characteristics of croplands across the African
continent region, such as regional variation in the sizes and numbers of
fields. We demonstrate this by analyzing these properties at two scales.
The first was a country scale assessment, in which we calculated the
median size and site-level number of fields for each country having at
least 30 labelled sites. We also examined annual trends in field size
for countries that had at least 30 sampled sites in each of the seven
years. In the second analysis, we calculated and mapped the median field
size and numbers within all 1° grid cells across the sampled region that
contained at least 10 digitized sites. To calculate field size and
numbers, we first filtered the full set of assignments by selecting the
highest quality assignment for each multi-assignment site, selecting
only those sites that had at least 1 field digitized. To estimate field
size, we extracted the polygons that were contained entirely within the
0.005° target polygon, in order to avoid biases caused by the truncation
of fields overlapping the boundary, transformed the polygons into an
Albers Equal Area Conic projection, and then calculated the area in
hectares for each polygon. For field number, we counted the total
polygons for each site. At each scale of analysis, we used the median
rather than the mean to avoid the influence of outliers, which could
arise from either digitization error or the chance that the sample of
sites was biased towards one side of the size class/number distribution.
To evaluate annual country-level trends in median field size, we used
the R package \texttt{mblm} (Komsta 2019) to apply Siegel's repeated
median regression (Siegel 1982), a modified version of Theil-Sen
regression (Theil 1950, Sen 1968), which is robust to outliers. We used
the Kendall test to assess the association between year and field size
(Sen 1968, Hollander and Wolfe 1973).

\section{Results}\label{results}

\subsection{Total assignments}\label{total-assignments}

\begin{figure}

\centering{

\pandocbounded{\includegraphics[keepaspectratio]{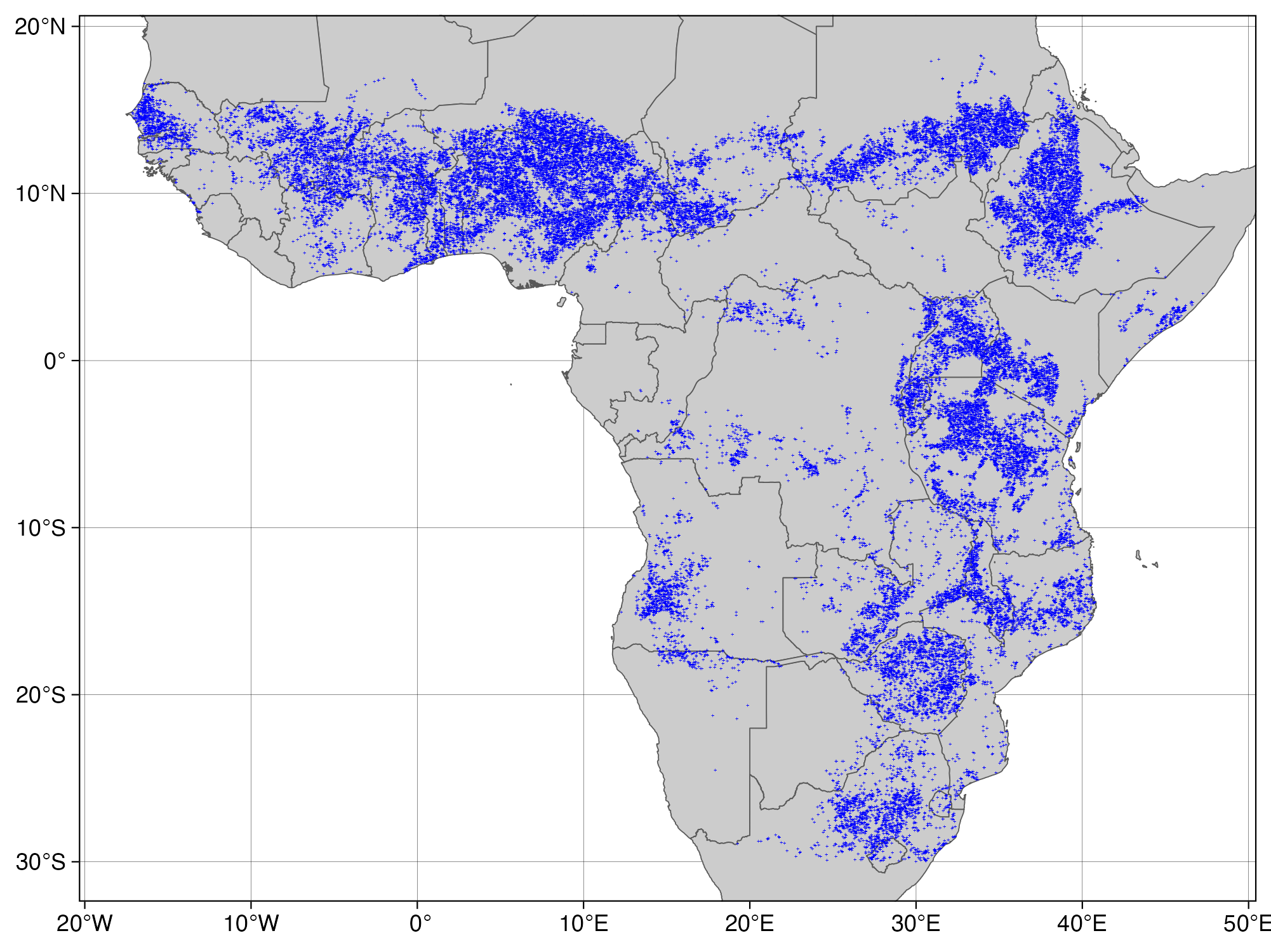}}

}

\caption{\label{fig-assnmap}The locations of the 33,746 sites that were
labelled, indicated by blue crosses.}

\end{figure}

The main labelling tasks (Classes 2 and 4) were undertaken between
November, 2023 and completed in mid-March, 2024. A total of 42403
assignments from 33746 unique sites Figure~\ref{fig-assnmap} were
completed, broken down into the classes detailed in
Table~\ref{tbl-assns}. Class 1 assignments were divided into 4
sub-categories (Classes 1a-1d). Class 1a were a subset of labels that
were assessed to be of passing quality by an initial independent review.
These were used to provide the reference labels used for assessing Q
assignments on the platform. Class 1b comprised the remaining Class 1
assignments completed by the expert teams. Class 1c were the total
number of Q assignments completed by the main labelling teams,
i.e.~those scored against the Class 1a labels. Finally Class 1d
assignments arose from the Class 1b sites that were re-mapped by 1-3
members of the primary labelling teams, to provide additional
assignments for comparison with the Class 1b labels, and for assessing
label risk along with the Class 4 labels. A total of 825,395 polygons
were digitized across the various labelling assignments.

\begin{longtable}[]{@{}lr@{}}

\caption{\label{tbl-assns}Number of assignments mapped per Class.}

\tabularnewline

\toprule\noalign{}
Class & n \\
\midrule\noalign{}
\endhead
\bottomrule\noalign{}
\endlastfoot
1a & 797 \\
1b & 1376 \\
1c & 2598 \\
1d & 2433 \\
2 & 32167 \\
4 & 3032 \\

\end{longtable}

Alongside information on the label class, several additional variables
were collected for each labelling assignment (see
Table~\ref{tbl-catalog} in the Appendix), such as the time spent
collecting the label, a set of quality metrics, and the number of
digitized fields.

\subsection{Label quality and risk}\label{label-quality-and-risk}

The overall quality of the labels was assessed using
Equation~\ref{eq-score} and the label reviews, and summarized in
Table~\ref{tbl-scores}. Summary scores include the overall mean Qscore,
its three main components N, Edge, Area, as well as Rscore, the mean of
the binary expert review scores, which represents the proportion of
passing assignments (0.7) out of the 2999 reviewed assignments. Of
these, 1787 were given the highest ratings (3-4) according to the
4-class review score. Of the four measures calculated from the quality
control algorithm Equation~\ref{eq-score}, the average of the overall
Qscore and Area were highest (0.61 and 0.75, respectively), while Edge
was the lowest and N the second lowest.

\begin{longtable}[]{@{}rrrrr@{}}

\caption{\label{tbl-scores}The average scores for the various quality
metrics assessed during the project. Qscore is the weighted mean of N,
Edge, and Area, which were calculated using the platform's quality
control algorithm, while Rscore is the overall proportion of
expert-reviewed assignments assessed as passing.}

\tabularnewline

\toprule\noalign{}
Qscore & N & Edge & Area & Rscore \\
\midrule\noalign{}
\endhead
\bottomrule\noalign{}
\endlastfoot
0.61 & 0.33 & 0.05 & 0.75 & 0.7 \\

\end{longtable}

\begin{figure}

\centering{

\pandocbounded{\includegraphics[keepaspectratio]{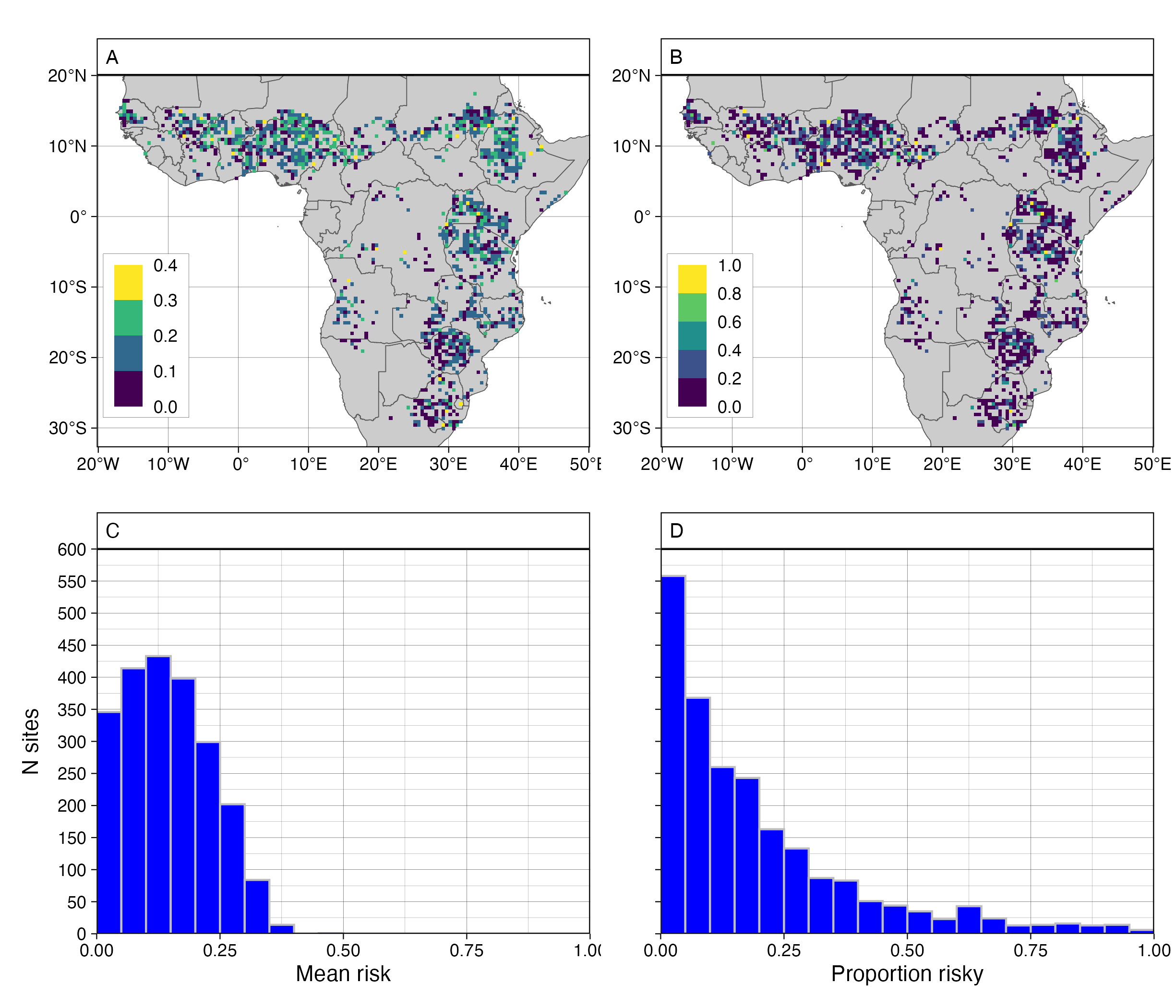}}

}

\caption{\label{fig-risk}Label risk, a measure of label uncertainty for
sites mapped by three or more labellers. Panels A and C show the mean
risk per pixel, mapped into 0.5° pixels and shown as a histogram, while
panels B and D relate to the proportions of each site covered by risky
pixels (r\textgreater0.34).}

\end{figure}

The average site-level label risk across the 2191 that were mapped by 3
or more labellers was 0.145, with the distributions of risk illustrated
in Figure~\ref{fig-risk}. To visualize the risk metrics across the
region, we average both mean label risk and the proportion of risk
labels within a 0.5° grid (Figure~\ref{fig-risk} A,B), which reveals a
higher occurrence of risky labels along the entire sub-Sahelian region
(from Ethiopia to the northwestern coast). Overall the mean risk for
half of the labels was less than 0.14 and less than 0.3 for 95\% of
labels (Figure~\ref{fig-risk} C). Beyond this, half of sites had
\(\leq\) 13\% of their areas covered by risky pixels
(risk\textgreater0.34), while 90\% had less than 48\% risk pixels
(Figure~\ref{fig-risk} D).

\subsection{Cropland characteristics}\label{cropland-characteristics-1}

\begin{figure}

\centering{

\pandocbounded{\includegraphics[keepaspectratio]{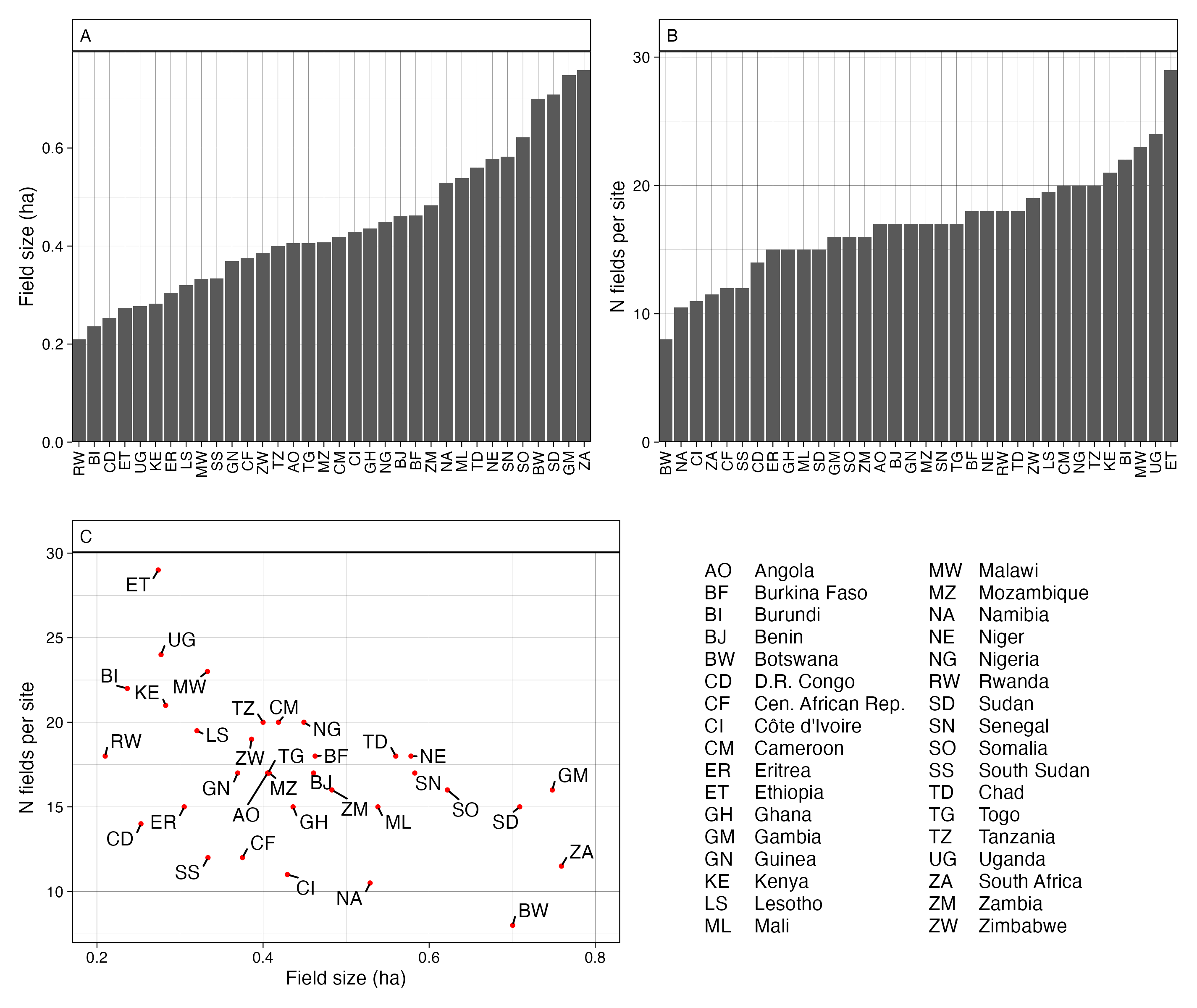}}

}

\caption{\label{fig-fsizect}The country-scale median field size (A) and
number of fields per sample site (B), as well as a scatter-plot of the
relationship between the two measures (C).}

\end{figure}

The analysis of median field size and count at the country scale shows
that the largest fields are in South Africa, Gambia, Sudan, and
Botswana, and the smallest are in Rwanda, Burundi, the Democratic
Republic of Congo, and Ethiopia Figure~\ref{fig-fsizect}. The median
number of fields per site in each country is highest in Burundi, Malawi,
Uganda, and Ethiopia, and lowest in Botswana, Namibia, Namibia, and
South Africa (Figure~\ref{fig-fsizect} B). In general, there is an
inverse relationship between field size and number
(Figure~\ref{fig-fsizect} C), with some notable exceptions, such as in
the Democratic Republic of Congo, where the median field size and number
are both small.

The assessment of trend in field sizes showed slopes that were both
positive and negative (Figure~\ref{fig-sizebyyear}). However, the only
significant relationships (p \textless{} 0.05) were the positive
associations between year and field size in Tanzania and Chad,
indicating a potential increases in field size between 2017-2023.

\begin{figure}

\centering{

\pandocbounded{\includegraphics[keepaspectratio]{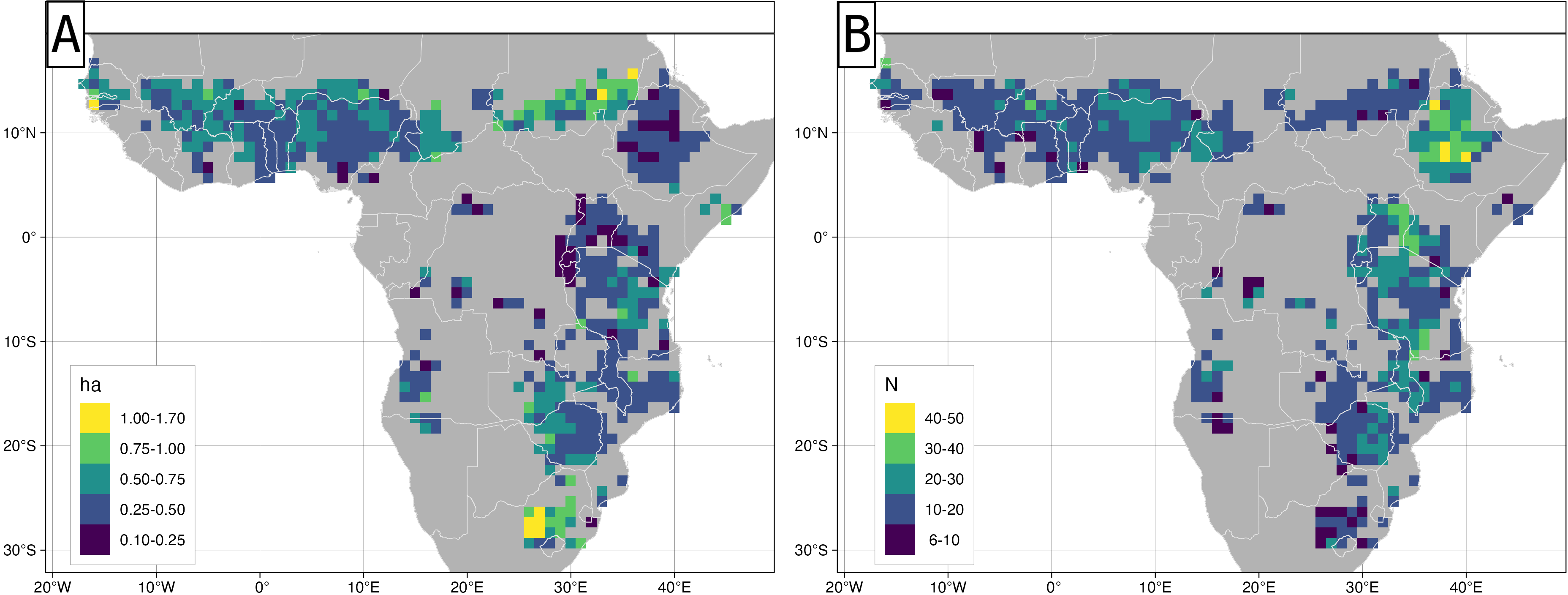}}

}

\caption{\label{fig-sizedensmap}Maps of median size (A) and number (B)
for each 1 degree cell having more than 10 sampled sites per cell.}

\end{figure}

The gridded assessment provides a spatial view of field size and count
across the region (Figure~\ref{fig-sizedensmap}) that reinforces the
general patterns revealed in the country-scale assessment (e.g.~larger
field sizes and smaller field counts in both South Africa and Sudan),
while providing additional insight into how fields vary within
countries. For example, fields in Botswana are confined to a small
number of cells near the country's southern and eastern boundaries
(Figure~\ref{fig-sizedensmap} A). Field sizes appear to be larger in the
east and smaller in the west of Tanzania, while the opposite patterns
are evident in Zimbabwe and Ghana. Ethiopia has uniformly small field
sizes (nearly all cells have median sizes \textless0.5 ha), while field
counts (Figure~\ref{fig-sizedensmap} B) are generally high but vary from
6-10 fields per site to 40-50 per site, which are the highest mapped
field counts across the entire region.

\section{Discussion}\label{discussion}

This dataset provides an extensive sample of field boundaries collected
from the years 2017-2023 at 33746 unique sites across Africa, from south
of the Sahara to the northern Karoo Desert (30° S). The labels are
provided along with the high resolution Planet image chips on which they
were digitized, and a set of quality metrics that can be used to
understand and control for the effects of label error. In addition to
these metrics, additional insight into label uncertainty (e.g.~risk,
Figure~\ref{fig-risk}) can be derived from the subset of sites that were
mapped by multiple labellers.

\subsection{Potential uses}\label{potential-uses}

These labels can be used to train and assess a variety of machine
learning models intended to map agricultural systems. The dataset was
collected for the primary purpose of developing boundary-aware semantic
segmentation models, with the ultimate goal of extracting field
instances (e.g. Wang et al. 2022, Lu et al. 2024, Rufin et al. 2024). An
example of such an application is in Khallaghi et al. (2024), who
converted field polygons collected on Planet imagery into the same
3-class labels as those illustrated in Figure~\ref{fig-label}, and used
them to train a modified U-Net (Ronneberger et al. 2015) to classify
field interiors, field edges, and non-field areas. Similarly, these data
could be adapted to train a multi-task learner, such as the ResUNet-a
developed by Waldner and Diakogiannis (2020) and Diakogiannis et al.
(2020) , which was simultaneously trained to predict field extent, field
edge, the distance from inside a field to its nearest edge, and to
reconstruct the original input image. The labels could also provide a
basis for transferring such models pre-trained on the substantially
larger, but slightly coarsely resolved (10 m) Fields of the World
dataset (Kerner et al. 2024), which may also enable higher
edge-delineation performance.

Beyond those particular applications, these data could also be converted
to crop/non-crop labels for training binary semantic segmentation
models, or could provide point samples for developing models that do not
require spatial context (e.g. Xiong et al. 2017). The quality measures
may also be used to understand and control for the effects of label
error on model performance. Lower quality labels can be used strictly
for training neural networks, which have some robustness to label error,
while the highest quality labels could be divided into subsets used for
model fine-tuning and final validation (Burke et al. 2021). To more
explicitly test the sensitivity of models to training errors (Elmes et
al. 2020), low and high quality subsets could be created to train
separate models that are then assessed against a common high quality
subset (e.g. Estes et al. 2022). The metrics could also be incorporated
directly into model loss functions that apply variable weighting based
on quality (e.g. Song et al. 2023).

\begin{figure}

\centering{

\pandocbounded{\includegraphics[keepaspectratio]{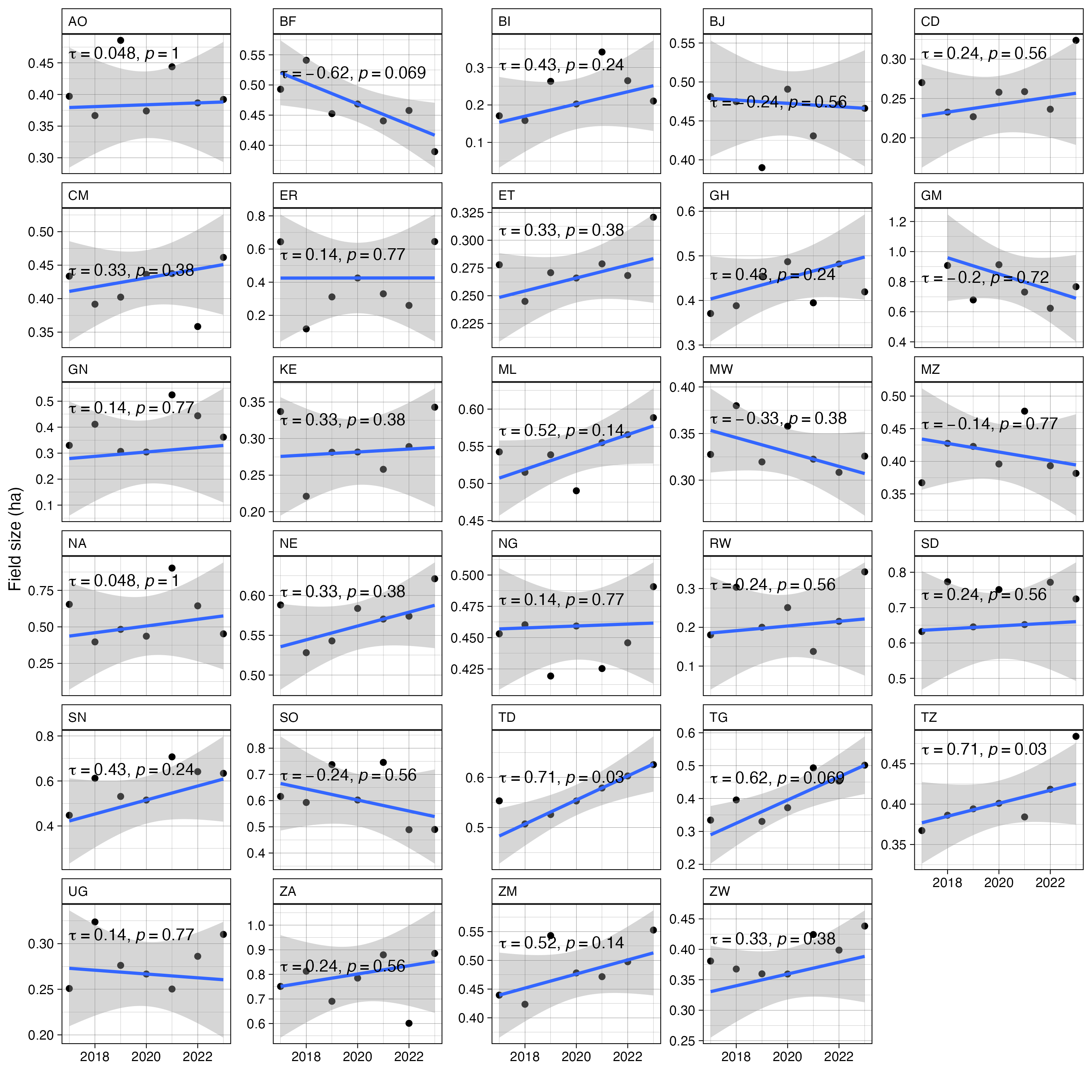}}

}

\caption{\label{fig-sizebyyear}The median field size by year for each
country having at least 30 labelled sample per year. The slope of the
Siegel regression and confidence interval is shown for each series,
along with Kendall's Tau coefficient and the p-value from its
significance test.}

\end{figure}

Besides modeling applications, the labels on their own provide
information on the characteristics of croplands across the majority of
Africa's arable agricultural regions. This includes insight into
regional variations in field size, which updates the African portion of
the field size estimates developed by Lesiv et al. (2019), along with
additional information on field density. Other potential applications
include estimating farm size from field size (e.g. Jänicke et al. 2024),
which may help to identify changes in farm ownership and management
patterns, such as the growth of medium-scale farming enterprises driven
by urban middle-class investors, which has been identified in earlier
studies based on household survey data (Sitko and Jayne 2014, Sitko and
Chamberlin 2015, Jayne et al. 2016). For example, the temporal analysis
presented here here suggests a trend towards increasing field sizes in
Tanzania Figure~\ref{fig-sizebyyear}, which could represent a
continuation of the growth in medium size farms observed between
2008-2012 (Jayne et al. 2016). Additional information would be needed to
verify this possibility, given the spatio-temporal disconnect in the
labels and their relatively brief, seven-year record, but the patterns
suggested by this dataset may help target follow-up work.

\subsection{Limitations}\label{limitations}

This dataset has several limitations that can affect its usefulness for
modeling applications, as well as its ability to represent agricultural
characteristics. The first of these is the inherent problem of labelling
error, which is influenced by a variety of factors, such as the
frequency and precision of noding (e.g. Gong et al. 1995) and
interpretation differences, which can vary substantially between
observers (McRoberts et al. 2018, Foody 2024). All of these factors, but
especially the latter, affect the quality of these labels to varying
degrees, as indicated by the accompanying quality metrics. A major
reason for interpretation error relates to the original 4.88 m effective
resolution of the Planet imagery. Although we resampled these imagery to
\textless3 m, which helped to slightly improve the edge visibility, the
resolution is still too coarse to effectively distinguish the smallest
fields, particularly in high density croplands, which is reflected by
the low values of the field count and edge accuracy metrics
(Table~\ref{tbl-scores}). To most effectively reduce such interpretation
error, field boundaries should be delineated on images that have
\textless2 m resolution (Wang et al. 2022). This limitation means that
the quality and representativeness of these data are biased towards
larger size classes. Field delineation models trained using these data
should not be expected to achieve high performance for fields \textless1
ha (Wang et al. 2022).

A second major limitation of this dataset is the separation of its
temporal and spatial domains, which makes it harder to assess the
consistency of model performance across years. We did not collect
repeated labels at the same sites because the combined interpretation
and digitization error would likely confound multi-year model
assessments, and because our primary goal was to broaden the geographic
scope of the sample. We increased the temporal variability of our sample
by randomizing the year of imagery to be labelled, which should improve
the temporal consistency of models trained with these data. This
consistency can be further enhanced by model-specific techniques such as
dropout regularization or photometric augmentation (e.g. Yaras et al.
2021, Khallaghi et al. 2024). However, to fully assess the temporal
consistency of any model, a separate set of site-specific, multi-year
reference labels should be developed.

\section*{Data availability}\label{data-availability}
\addcontentsline{toc}{section}{Data availability}

The vectorized labels and image chips for this dataset are publicly
available on the Registry of Open Data on Amazon Web Services\footnote{\url{https://registry.opendata.aws/africa-field-boundary-labels/}}
and on Zenodo\footnote{\url{https://zenodo.org/records/11060871}}. These
links and the code used to process and analyze these data, along with
the source files for this manuscript, are available on
GitHub\footnote{\url{https://github.com/agroimpacts/lacunalabels}}. The
label data and associated images may be used in accordance with Planet's
participant license agreement for the NICFI contract\footnote{\url{https://assets.planet.com/docs/Planet_ParticipantLicenseAgreement_NICFI.pdf}}.

\section*{Acknowledgements}\label{acknowledgements}
\addcontentsline{toc}{section}{Acknowledgements}

The primary funding to support this research was provided by the Lacuna
Fund, with additional support provided by the National Science
Foundation (Awards \#1924309 and \#2439879). Support to develop the
labelling platform was provided by Omidyar Network's Property Rights
Initiative, now PLACE. We gratefully acknowledge credits provided by the
European Space Agency (ESA) and SentinelHub for use of the service,
through support provided by the ESA Network of Resources Initiative.

\section*{References}\label{references}
\addcontentsline{toc}{section}{References}

\phantomsection\label{refs}

\section*{Appendix}\label{appendix}
\addcontentsline{toc}{section}{Appendix}

\begin{longtable}[]{@{}
  >{\raggedright\arraybackslash}p{(\linewidth - 2\tabcolsep) * \real{0.3472}}
  >{\raggedright\arraybackslash}p{(\linewidth - 2\tabcolsep) * \real{0.6528}}@{}}
\caption{The names and descriptions of variables provided in the label
catalog.}\label{tbl-catalog}\tabularnewline
\toprule\noalign{}
\begin{minipage}[b]{\linewidth}\raggedright
Variable
\end{minipage} & \begin{minipage}[b]{\linewidth}\raggedright
Description
\end{minipage} \\
\midrule\noalign{}
\endfirsthead
\toprule\noalign{}
\begin{minipage}[b]{\linewidth}\raggedright
Variable
\end{minipage} & \begin{minipage}[b]{\linewidth}\raggedright
Description
\end{minipage} \\
\midrule\noalign{}
\endhead
\bottomrule\noalign{}
\endlastfoot
name & Unique site ID, prefixed with two character ISO country code \\
Class & Label class (1a-1d, 2, 4) \\
assignment\_id & Identifier for each unique mapping assignment (1
mapping by 1 labeller) \\
Labeller & Anonymous identifiers for each labeller \\
completion\_time & Date and time the labelling assignment was
completed \\
label\_time & Total time spent on the assignment \\
status & A system assigned value, including ``Rejected'' (failed Q
assignment), ``Untrusted'' (assignment completed during time when
labeller had low rolling average Q score); ``Approved'' assignment
passing the Q threshold, or non-Q assignment passed when labeller's
average Q score was above the quality threshold \\
Score & Weighted mean Quality score, comprised of N, Edge, Area, and
Categorical metrics derived from the Class 1a labels. \\
N & Agreement between number of fields mapped by a labeller and the
corresponding Class 1a labels \\
Edge & Nearness of labeller's field boundaries to those in a
corresponding Class 1a label set \\
Area & Agreement between a labeller's mapped area of fields and
non-fields and those of the corresponding Class 1a labels \\
FieldSkill, NoFieldSkill & A Bayesian metric of a labeller's skill in
mapping field and non-field areas, respectively (see Estes et al.
2022) \\
Categorical & Agreement in the label assigned to each polygon delineated
by the labeller and those in the Class 1a labels. \\
rscore & A 1-4 ranking assigned to a given assignment by a supervising
expert during independent review. Note: decimal values appear for cases
where the same assignment was assessed more than once by experts \\
rscore2 & A simplified binary version of \texttt{rscore}, where 0
indicates failing and 1 indicates passing. \\
Qscore & Each labeller's overall average Score, assigned as a general
confidence measure applied to all Class 2 and 4 assignments undertaken
by the labeller \\
QN & Each labeller's overall average N score, to provide a general
measure of each labeller's tendency to over or under-segment fields
(lower score typically mean under-segmentation). \\
Rscore & Each labeller's overall average rscore2, assigned as a second
generalized measure of confidence. Experts (except one) were assigned
the same measure, based on initial reviews conducted by a separate team
at Clark \\
x, y & The centroid of each site, in decimal degrees \\
farea & The average area (in ha) of fields digitized in each
assignment \\
nflds & The average number of fields digitized in each assignment \\
tile & The unique identifier of the 0.05° image tile containing the
labelling site \\
image\_date & The collection date of the image being labelled. The month
and day represent the central date of images from August, 2020 and
earlier, which were drawn from 6-month composites, while later images
were collected in the month indicated, with the 15th being the central
date for the month \\
chip & The chip identifier (a concatenation of name and image-date) \\
label & (In the demonstration catalog only). The identifier of
rasterized label chip, concatenating the name, image-date, and
assignment\_id. \\
\end{longtable}

\end{document}